\documentclass[11pt]{article}

\usepackage[preprint]{acl}
\usepackage{amssymb}   
\usepackage{times}
\usepackage{latexsym}
\usepackage[T1]{fontenc}
\usepackage[utf8]{inputenc}
\usepackage{microtype}
\usepackage{inconsolata}
\usepackage{graphicx}
\usepackage{amsmath}
\usepackage{subcaption}
\usepackage{xcolor}
\usepackage{booktabs}
\usepackage{colortbl}
\usepackage{multirow}
\usepackage{makecell}
\usepackage{textcomp}
\definecolor{mygray}{gray}{0.92}
\definecolor{ForestGreen}{RGB}{34,139,34}
\definecolor{lightgreen}{rgb}{0.886,0.941,0.851}
\newcommand{\fg}[1]{\textcolor{ForestGreen}{#1}}

\begin{document}

\title{DCP-Prune: Ultra-Low Token Pruning with Distribution Consistency Preservation}

\author{
    \textbf{Xifeng Xue\textsuperscript{1}},
    \textbf{Xiaokang Wang\textsuperscript{2}},
    \textbf{Zirui Li\textsuperscript{1}},
    \textbf{Ming-Ming Cheng\textsuperscript{1}},
    \textbf{Guolei Sun\textsuperscript{1}\thanks{Corresponding author.}}
    \\
    \textsuperscript{1}College of Computer Science, Nankai University, Tianjin, China\\
    \textsuperscript{2}Nanjing University of Posts and Telecommunications, Nanjing, China
}

\maketitle

\begin{abstract}
Recent vision token pruning methods effectively preserve model performance under moderate token budgets but become unstable under ultra-low token budget. Our analysis shows that as the pruning budget decreases, accuracy degradation is often accompanied by larger feature distribution shifts. Critically, the degree of this distribution shift strongly correlates with performance degradation. To better characterize this phenomenon, we introduce a lightweight distribution consistency metric to estimate the distribution shift between retained and full tokens. Motivated by these observations, we propose a two-stage pruning framework consisting of Anchor-Context Graph Recovery (ACGR) and Text-Aware Token Cluster Selection (TATCS). Specifically, ACGR transfers contextual information before token removal, while TATCS dynamically re-selects representative tokens when severe distribution shift is detected. Extensive experiments demonstrate that our method achieves superior and more stable performance under ultra-low token budget. Notably, it retains 92.1\% of the upper-bound average performance on LLaVA-1.5-7B with only 16 visual tokens. 
The code will be released at: \url{https://github.com/EMVision-NK/DCP-Prune}.
\end{abstract}

\section{Introduction}
Vision-language models (VLMs) have recently demonstrated strong performance across a broad range of multimodal tasks, including visual question answering \cite{dai2023instructblip, alayrac2022flamingo, huang2023language, li2023blip}, image understanding \cite{zhu2024minigpt, liu2023visual, chen2022pali,chen2024internvl}, and multimodal dialogue \citep{li2024llama,wang2024qwen2,li2024llavaonevision, wang2024cogvlm}. 
However, the computational cost of VLMs grows rapidly with increasingly complex visual inputs.
In VLMs, the number of visual tokens scales rapidly with higher image resolutions \citep{guo2024llava, dehghani2023patch, dong2024internlm}, multi-image inputs \citep{wang2023visionllm, ye2023mplug, moon2024anymal,chen2023shikra}, and video sequences \citep{li2024llavanextinterleave, arnab2021vivit, bain2021frozen, sun2019videobert}.
As a result, memory usage and inference latency increase substantially, posing a significant obstacle to practical deployment and making visual token pruning a critical research problem.

\begin{figure*}[t]
    \centering
    \includegraphics[width=1.0\linewidth,clip,trim=0.1cm 0.1cm 0.1cm 0.1cm]{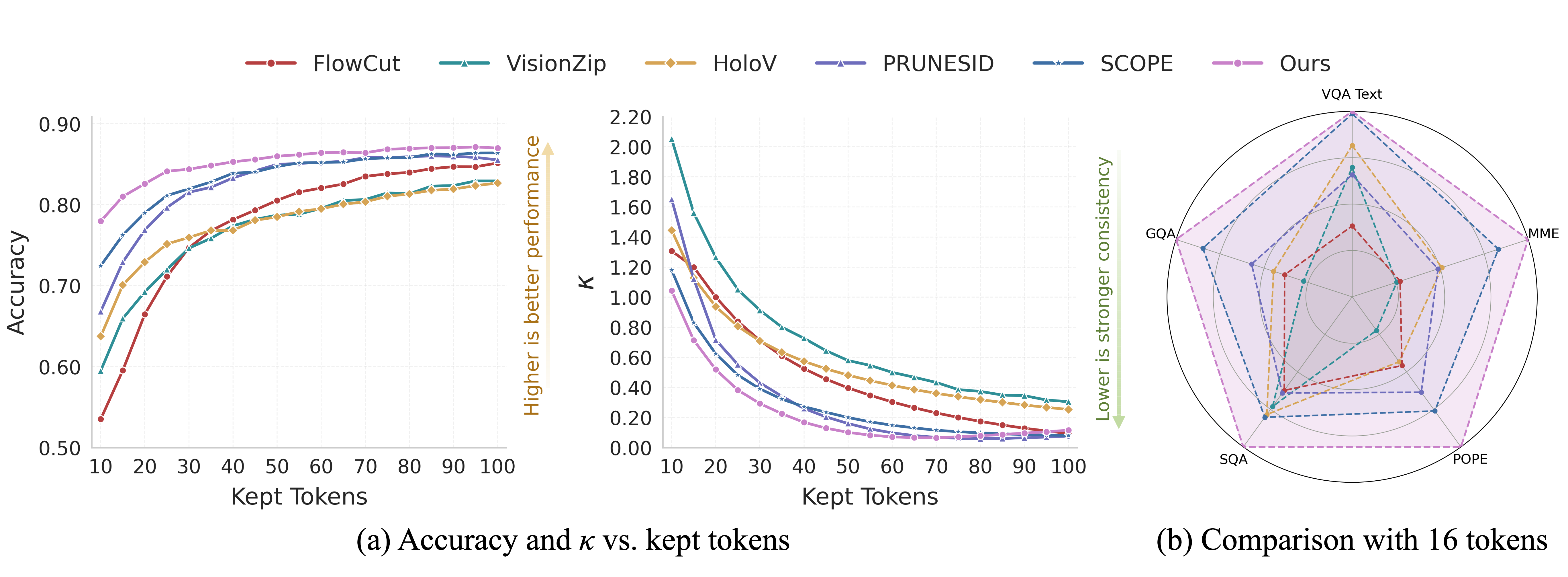}
    \vspace{-20pt}
    \caption{(a) Relationship between kept tokens, distribution deviation ($\kappa$), and accuracy for LLaVA-1.5-7B on POPE dataset. (b) Performance comparison under the ultra-low budget of 16 retained tokens.
}
    \vspace{-15pt}
    \label{fig:ackl}
\end{figure*}

Existing visual token pruning methods are generally categorized into importance-oriented methods and coverage-oriented methods, and both have shown strong performance under moderate token budgets \citep{jiang2025kind, zhang2025beyond, tan2025tokencarve, zhang2025vscan}. 
However, their performance often becomes unsatisfactory under \textit{ultra-low} token budgets.
Specifically, importance-oriented methods preserve salient tokens, but often suffer from redundancy and loss of contextual information under aggressive compression \citep{shang2025llava,sun2025lvpruning,yang2025vflowopt,li2025balanced}. 
In contrast, coverage-oriented methods prioritize spatial or semantic diversity, but may retain tokens that are less informative for reasoning \citep{kim2025training,chen2025ipcv, zhang2025trimtokenator, yu2026visiontrim}.
Consequently, under ultra-low token budget, both strategies progressively drive retained tokens away from the original feature distribution, degrading performance.

The distribution shift between retained and full tokens leads to performance degradation under ultra-low token budget \citep{zhang2026rcp, duan2022network}.
To better characterize this phenomenon, we introduce a lightweight distribution consistency metric:
\begin{equation}
\begin{aligned}
\mathcal{D}(X_{k},X_{f})&=
\frac{
\mathrm{Tr}(\Sigma_{k})+d\,\mathrm{Var}(\mu_{k})
}{
\mathrm{Tr}(\Sigma_{f})+d\,\mathrm{Var}(\mu_{f})
}, \\
\kappa&=|\mathcal{D}(X_{k},X_{f})-1|,
\end{aligned}
\label{eq:tatcs_instability}
\end{equation}
where $X_f$ and $X_k$ denote the full and kept token sets, respectively. $\mathrm{Tr}(\Sigma)$ measures feature dispersion, while $\mathrm{Var}(\mu)$ reflects shifts in channel-wise feature statistics.
Using this metric, Fig.~\ref{fig:ackl}(a) shows that reducing the token budget consistently enlarges the distributional discrepancy between retained and full tokens, accompanied by severe performance degradation. 
This aligned trend suggests that distributional discrepancy serves as an informative indicator of pruning regimes prone to performance degradation.

Motivated by the relation between distributional discrepancy and performance degradation, we design a two-stage framework for ultra-low visual token pruning.
During progressive pruning, \textbf{Anchor-Context Graph Recovery} (ACGR) transfers contextual information from tokens to be discarded into retained tokens, thereby alleviating loss of semantic information and maintaining token distribution under aggressive pruning.
At the final pruning stage, \textbf{Text-Aware Token Cluster Selection} (TATCS) is triggered under significant feature distribution shifts to reselect representative tokens with textual guidance, helping retain task-relevant visual information while improving feature distribution consistency.
As illustrated in Fig.~\ref{fig:ackl}(b), our framework achieves much better performance under ultra-low token budget, compared with existing methods.
Our contributions are threefold:

\textbf{(1)} We observe that ultra-low token pruning induces distribution shift between retained and full tokens, and further introduce a lightweight metric to quantify this shift, showing that the discrepancy closely aligns with performance degradation.

\textbf{(2)} We propose a two-stage pruning framework consisting of Anchor-Context Graph Recovery (ACGR) and Text-Aware Token Cluster Selection (TATCS) to respectively mitigate contextual information loss and distribution shift under ultra-low token budget.

\textbf{(3)} Extensive experiments on various VLMs demonstrate that our method consistently outperforms existing pruning methods under ultra-low token budget, retaining 92.1\% performance on LLaVA-1.5-7B with 16 tokens under (97.2\% reduction) and 91.9\% performance on LLaVA-NeXT-7B with 80 tokens under (97.2\% reduction).

\begin{figure*}[!htbp]
    \centering
    \includegraphics[width=1.0\linewidth]{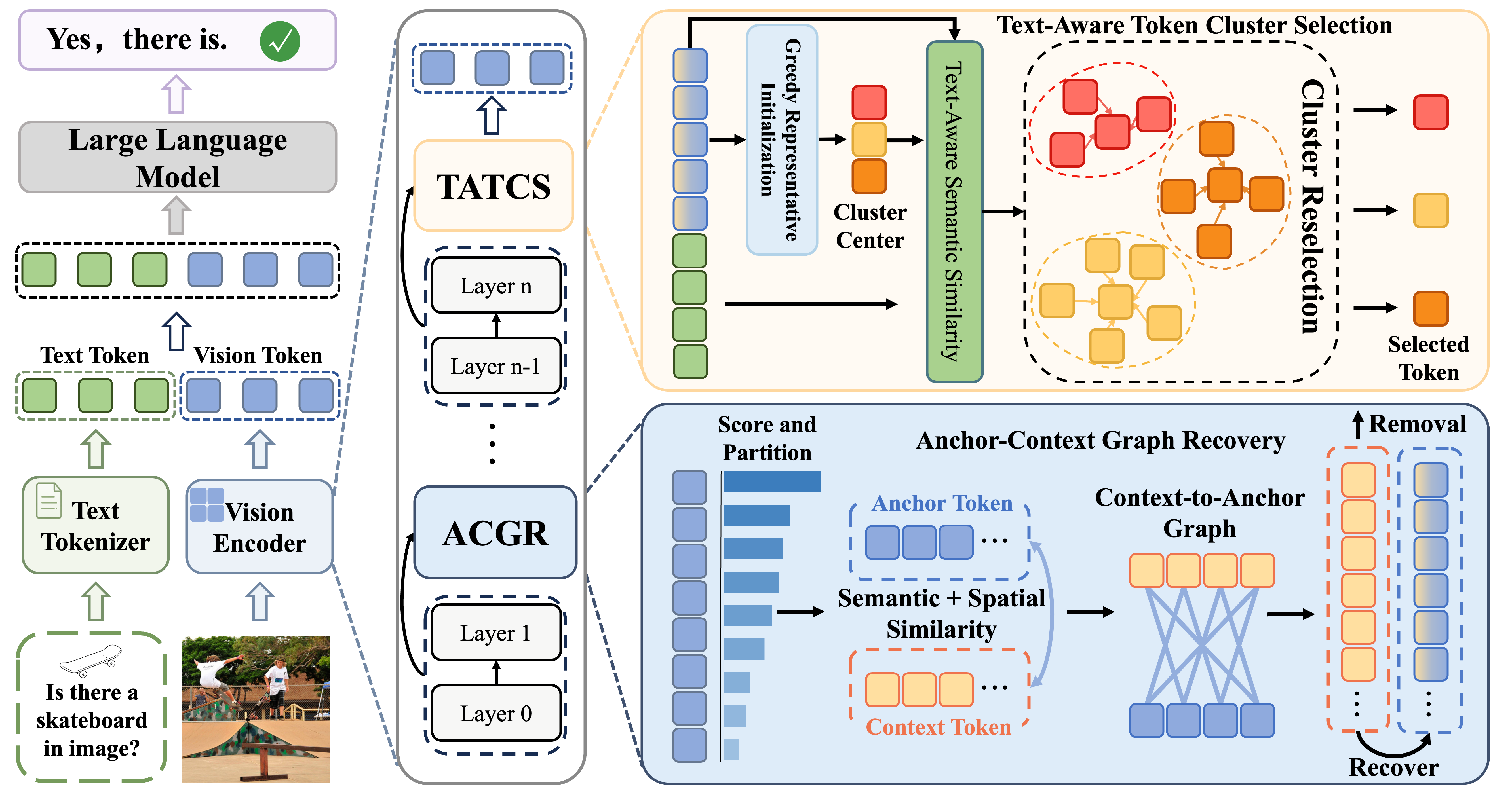}
    \vspace{-20pt}
    \caption{{Overview of the proposed two-stage ultra-low token pruning framework. ACGR transfers contextual information from pruned context tokens to retained anchor tokens to alleviate semantic loss. TATCS mitigates distribution drift by reselecting representative tokens using text-aware semantic similarity.
    }}
    \vspace{-15pt}
    \label{fig:overview}
\end{figure*}

\section{Related Works}

\paragraph{Visual Token Pruning in VLMs.}
Visual token pruning has emerged as an effective strategy for improving VLMs efficiency by removing redundant visual tokens during inference. 
Early studies show that many visual tokens contribute little to predictions and can be safely pruned based on attention or feature responses \citep{chen2024image}.
Subsequent work improves token selection by incorporating information flow or transformation dynamics, such as \citep{tong2025flowcut}. 
Other approaches attempt to balance token importance and diversity \citep{deng2025scope} or explicitly model redundancy and diversity in token selection \citep{fang2026prunesid}. 
Despite these advances, most methods select tokens by importance or coverage, largely overlooking distributional consistency. Under ultra-low token budgets, retained tokens can deviate from the original feature distribution, making it harder to preserve a compact yet representative subset.

\paragraph{Distribution Consistency in Token Pruning.}
Under ultra-low token pruning, retained visual token distributions progressively deviate from the original full token distribution.
Recent studies on visual token pruning show that aggressive compression introduces selection bias and weakens fine-grained visual understanding \citep{endo2025feather}.
More broadly, distribution modeling has been widely used to characterize representation discrepancy. For example, \citep{liang2025distribution} models semantic features with Gaussian distributions for feature alignment, while \citep{zhang2023properties} theoretically analyzes divergence measures between Gaussian feature distributions.
These studies suggest that feature distributions can be characterized through their statistical properties, providing a way to assess representation mismatch. 
In contrast, most existing visual token pruning methods focus on token importance or diversity, without explicitly considering whether the retained subset preserves the original statistical structure. Under aggressive pruning, this mismatch becomes more pronounced, making the retained tokens less representative and degrading multimodal reasoning performance.

\section{Method}

\subsection{Overview}

Given a set of $d$-dimensional visual tokens $X_f \in \mathbb{R}^{N \times d}$, our objective is to select a compact kept subset $X_k \in \mathbb{R}^{K \times d}$ under an ultra-low token budget ($K \ll N$). 
During progressive pruning, the retained tokens should preserve task-relevant visual semantics while retaining complementary contextual information.
Meanwhile, the kept tokens representations should remain consistent with the original full token distribution under ultra-low token budget.

In this paper, we propose a two-stage pruning framework, illustrated in Fig.~\ref{fig:overview}. During progressive pruning, ACGR is employed to aggregate contextual information from discarded tokens into retained anchor tokens prior to token removal, thereby preserving richer visual semantics under aggressive compression. At the final pruning stage, where severe distribution shift may arise, TATCS is introduced to perform text-aware representative token reselection from the original token pool, helping maintain feature distribution consistency under ultra-low token budget.

\subsection{Anchor-Context Graph Recovery}
\label{acgr}

Under ultra-low token budget, directly discarding visual tokens inevitably removes complementary semantic and spatial information. 
As pruning intensifies, the contextual relationships among tokens are increasingly disrupted, making the remaining tokens less representative of the original visual semantics.
To address this issue, we propose Anchor-Context Graph Recovery (ACGR), which transfers contextual information from discarded contextual tokens to retained anchor tokens before pruning, thereby preserving richer visual semantics under aggressive pruning.

Given the token set $X=\{x_i\}_{i=1}^{M}$ at the current pruning stage, we first perform \textit{score and partition} by computing an overall importance score for each token as:
\begin{equation}
\label{acgr:score}
s_i=s_i^{\text{cls}}+s_i^{\text{cos}}+s_i^{\text{var}}+s_i^{\text{abl}}.
\end{equation}
Here, $s_i^{\text{cls}}$, $s_i^{\text{cos}}$, $s_i^{\text{var}}$, and $s_i^{\text{abl}}$ measure CLS relevance, feature non-redundancy, transformation variation, and attention-ablation impact, respectively. 
These criteria jointly capture global semantic importance, redundancy suppression, feature dynamics, and token influence through attention interactions. Detailed formulations and normalization strategies are provided in Appendix~\ref{sec:score}.
Based on the resulting scores $s_i$, the token set $X$ is then partitioned into an anchor set $X_a=\{x_j^a\}_{j=1}^{M_a}$ and a context set $X_c=\{x_i^c\}_{i=1}^{M_c}$, where $M_a$ and $M_c$ denote the numbers of retained anchor tokens and discarded context tokens, respectively.

Instead of directly discarding context tokens, ACGR constructs a semantic-spatial context-to-anchor graph to transfer complementary information from $X_c$ to $X_a$. 
Since semantically similar and spatially nearby tokens are more likely to provide complementary information, we measure token affinity using both \textit{semantic and spatial similarity}.
Accordingly, the edge weight between context token $x_i^c$ and anchor token $x_j^a$ is defined as
\begin{equation}
e_{ij}=\cos(x_i^c,x_j^a)\cdot
\exp\left(
-\frac{\|p_i^c-p_j^a\|_2^2}{\tau}
\right).
\end{equation}
where $p_i^c$ and $p_j^a$ denote token grid coordinates, and $\tau$ controls spatial decay.
To suppress noisy information transfer, each anchor token retains only its top-$k_n$ context neighbors. Finally, the retained anchor tokens are updated through residual aggregation over the \textit{context-to-anchor graph}:
\begin{equation}
\tilde{x}_j^a
=
x_j^a
+
\alpha
\sum_{i \in \mathcal{N}_{k_n}(j)}
\bar{e}_{ij} x_i^c,
\end{equation}
where
$\bar{e}_{ij}=
e_{ij}/\sum_{i\in \mathcal{N}_{k_n}(j)} e_{ij}$ denotes the normalized edge weight, $\mathcal{N}_{k_n}(j)$ denotes the top-$k_n$ neighbors of $x_j^a$, and $\alpha$ is the fusion coefficient.

The recovered anchor set $\tilde{X}_a=\{\tilde{x}_j^a\}_{j=1}^{M_a}$ maintains the same token budget as direct pruning while incorporating complementary semantic-spatial information from discarded tokens. As a result, ACGR mitigates the representation degradation caused by progressive pruning under aggressive compression.

\subsection{Text-Aware Token Cluster Selection}
\label{tatcs}

Although ACGR alleviates contextual information loss during progressive pruning, the retained token subset can still gradually drift from the original feature distribution under ultra-low token budget, reducing the representativeness of the kept tokens. 
To address this, TATCS is activated at the final pruning stage when significant distribution shift is detected, performing text-aware token reselection from $X$ to restore distributional consistency.

At the final pruning stage, we first compute the distribution consistency metric $\kappa$ using Eq.~\ref{eq:tatcs_instability} between the retained anchor tokens $X_a$ and the input token set $X$.
When $\kappa > \delta$, where $\delta$ is a predefined instability threshold, the retained subset is considered to exhibit severe distribution drift and becomes less representative of the original visual semantics. 
Under this condition, TATCS is applied at the final pruning stage to perform text-aware token reselection and restore feature distribution consistency.
A statistical interpretation of $\kappa$ is provided in Appendix~\ref{app:gaussian_kappa}.

Then, under the target token budget $K$, we perform \textit{greedy representative initialization} on $X$ to obtain a representative token subset.
Specifically, during each greedy selection step, every candidate token $x_i \in U$ is evaluated by
\begin{equation}
G_i=
\frac{1}{|U|}
\sum_{x_j\in U}
\cos(x_i,x_j)
-
\max_{x_p\in S}\cos(x_i,x_p)
+\hat{a}^{\text{cls}}_i,
\end{equation}
where $S$ and $U$ denote the selected token set and the unselected candidate token set, respectively, with $U=X$ at initialization, and $\hat a_i^{\mathrm{cls}}$ is the normalized CLS attention score.
At each iteration, the token with the largest gain $G_i$ is added to $S$. The process is repeated until $K$ representative tokens are selected, and the selected token set $S=\{z_k\}_{k=1}^{K}$ is used as the initial cluster centers.

Given the text feature $t$ extracted from the input text, we define a \textit{text-aware semantic similarity} between token $x_i$ and cluster center $z_k$ by jointly modeling their visual similarity and text-alignment consistency:
\begin{equation}
\label{tatcs:text}
\mathrm{sim}_{ik}
=
\cos(x_i,z_k)
-
\left|\cos(x_i,t)-\cos(z_k,t)\right|,
\end{equation}
where the first term captures visual similarity, while the second term enforces consistency in text relevance between tokens assigned to the same cluster.
Based on this similarity, each token is assigned through text-aware assignment:
\begin{equation}
c_i=\arg\max_k \mathrm{sim}_{ik}.
\end{equation}

The cluster centers are then iteratively updated by averaging the assigned token features:
\begin{equation}
\tilde{z}_k=
\frac{
\sum_{i=1}^{M}\mathbf{I}[c_i=k]x_i
}{
\sum_{i=1}^{M}\mathbf{I}[c_i=k]
},
\end{equation}
where \(\mathbf{I}[\cdot]\) is the indicator function and $\tilde{z}_k$ denotes the iteratively updated cluster center.

After iterative refinement, we perform \textit{cluster reselection} by selecting representative tokens as $
x_k^{*} = x_{i^{*}}$, 
$i^{*} = \arg\max_{i:c_i=k} \mathrm{sim}_{ik}$,  
to obtain the final compact token set $X_{\text{out}} = \{x_k^{*}\}_{k=1}^{K}$.
By applying text-aware token cluster selection only when significant distribution deviation occurs, TATCS maintains feature distribution consistency while improving the semantic representativeness of retained tokens under ultra-low token budget.

\section{Experiments}

\subsection{Experiment Settings}

\paragraph{Models.} We evaluate the effectiveness of our proposed token pruning strategy under ultra-low token budget on a set of representative vision-language models, including LLaVA-1.5 \citep{liu2024improvedllava}, LLaVA-NeXT \citep{liu2024llavanext}, Video-LLaVA \citep{lin2024video}, and Qwen2-VL \citep{wang2024qwen2}, covering both image and video understanding tasks.
Specifically, image understanding is evaluated on five widely used benchmarks: GQA \citep{hudson2019gqa}, MME \citep{fu2023mme}, POPE \citep{li2023evaluating}, SQA \citep{lu2022learn}, and TextVQA \citep{singh2019towards}, while video understanding is evaluated on TGIF \citep{jang2017tgif}, MSVD \citep{chen2011collecting}, and MSRVTT \citep{xu2016msr}. 
We compare against a set of recent competitive methods, including VisionZip \citep{yang2025visionzip}, HoloV \citep{zou2025don}, FlowCut \citep{tong2025flowcut}, PRUNESID \citep{fang2026prunesid}, and SCOPE \citep{deng2025scope}.

\paragraph{Implementation Details.} 
Our method is entirely training-free and operates exclusively at inference time.
For ACGR, pruning is applied every two Transformer layers in the vision encoder, progressively and uniformly reducing the token budget from the initial visual token number \(N\) to the target budget \(K\), while using \(k_n = 5\), a spatial decay factor \(\tau = 10\), and a fusion coefficient \(\alpha = 0.1\) across all experiments.
In TATCS, which is only potentially activated at the final pruning layer, re-selection is triggered when the threshold $\delta = 0.1$ is exceeded. 
Meanwhile, text features are extracted either from the text encoder aligned with the visual embedding space, when available \citep{radford2021learning}, or otherwise from the aligned LLM text representations.
All methods are evaluated under identical backbone and inference settings to ensure a fair comparison.

\subsection{Main Results}

\paragraph{Results on LLaVA 1.5.} 

Table~\ref{tab:llava1.57b} reports the performance of different methods on LLaVA-1.5-7B under varying token budgets. As the token budget decreases, the performance advantage of our method becomes progressively more pronounced. Specifically, our method achieves 96.3\% accuracy at 64 tokens, surpassing the strongest method by 0.5\%. The gain further increases to 0.7\% at 32 tokens and reaches 2.7\% at 16 tokens. These results highlight the effectiveness and robustness of our method, particularly in ultra-low token budget.

Moreover, existing methods degrade more rapidly as the token budget becomes more restrictive, whereas our method remains more stable.
When the token budget is reduced from 64 to 16, the accuracy of SCOPE declines from 95.8\% to 89.4\%, while our method only drops from 96.3\% to 92.1\%. 
Even more pronounced degradation is observed in FlowCut, whose accuracies decrease from 94.4\% to 82.3\%. These results suggest that our method is more robust under ultra-low token pruning and better preserves representation quality.

Furthermore, under the ultra-low token budget of 16 tokens, our method achieves the best performance across all evaluated benchmarks. 
In particular, it reaches 81.5\% on POPE, outperforming SCOPE by 4.6\%, and achieves 55.1\% on GQA, which is also higher than SCOPE at 53.4\%. 
These results show that our method exhibits slower performance degradation as the token budget becomes more aggressive. 
Overall, the results demonstrate that our method consistently maintains stronger performance than existing approaches under ultra-low token budget.

\begin{table}[t]
  \centering
  \small
  \setlength{\tabcolsep}{5pt}
  \renewcommand{\arraystretch}{1.05}
  \caption{Performance comparison under different visual token settings on LLaVA-1.5-7B. The vanilla number of visual tokens is 576, and the last column reports the average accuracy proportion w.r.t. the vanilla. The entries marked with $^{\dagger}$ are reported from the corresponding original papers.}
  \label{tab:llava1.57b}
  \resizebox{\columnwidth}{!}{
  \begin{tabular}{lcccccc}
    \toprule[1.2pt]
    \textbf{Method} & \textbf{GQA} & \textbf{MME} & \textbf{POPE} & \textbf{SQA} & \textbf{TextVQA} & \textbf{Avg.} \\
    \midrule
    \rowcolor{mygray}
    \multicolumn{7}{c}{\textit{Upper Bound, 576 Tokens} \ \textbf{(100.0\%)}} \\
    \textcolor{gray}{Vanilla} & \textcolor{gray}{61.9} & \textcolor{gray}{1862} & \textcolor{gray}{85.9} & \textcolor{gray}{69.5} & \textcolor{gray}{58.2} & \textcolor{gray}{100.0\%} \\
    \midrule
    \rowcolor{mygray}
    \multicolumn{7}{c}{\textit{Retain 64 Tokens} \ \fg{$(\downarrow 88.9\%)$}} \\
    VisionZip & 55.1$^{\dagger}$ & 1690$^{\dagger}$ & 80.9$^{\dagger}$ & 69.0$^{\dagger}$ & 55.5$^{\dagger}$ & 93.7\% \\
    HoloV     & 55.3$^{\dagger}$ & 1715$^{\dagger}$ & 80.3$^{\dagger}$ & \textbf{69.5}$^{\dagger}$ & 55.4$^{\dagger}$ & 94.0\% \\
    FlowCut   & 55.6$^{\dagger}$ & \textbf{1744}$^{\dagger}$ & 80.4$^{\dagger}$ & 69.1$^{\dagger}$ & 55.6$^{\dagger}$ & 94.4\% \\
    PRUNESID  & 57.1$^{\dagger}$ & 1733$^{\dagger}$ & 83.8$^{\dagger}$ & 67.8$^{\dagger}$ & 54.2$^{\dagger}$ & 94.7\% \\
    SCOPE     & 58.3$^{\dagger}$ & 1698$^{\dagger}$ & 83.9$^{\dagger}$ & 68.6$^{\dagger}$ & \textbf{56.6}$^{\dagger}$ & 95.8\% \\
    \rowcolor{lightgreen!80}
    Ours & \textbf{58.5}\hphantom{$^{\dagger}$} & 1719\hphantom{$^{\dagger}$} & \textbf{86.5}\hphantom{$^{\dagger}$} & 68.3\hphantom{$^{\dagger}$} & 55.7\hphantom{$^{\dagger}$} & \textbf{96.3\%} \\
    \midrule
    \rowcolor{mygray}
    \multicolumn{7}{c}{\textit{Retain 32 Tokens} \ \fg{$(\downarrow 94.4\%)$}} \\
    VisionZip & 51.9 & 1581 & 74.7 & 68.8 & 53.1 & 89.2\% \\
    HoloV     & 52.8 & 1598 & 76.1 & \textbf{68.8} & 53.7 & 90.2\% \\
    FlowCut   & 51.9 & 1558 & 75.5 & 68.3 & 52.6 & 88.8\% \\
    PRUNESID  & 54.8 & 1592 & 82.1 & 67.8 & 52.5 & 91.5\% \\
    SCOPE     & 56.4 & \textbf{1655} & 82.4 & 68.4 & \textbf{54.8} & 93.7\% \\
    \rowcolor{lightgreen!80}
    Ours & \textbf{57.0} & 1654 & \textbf{85.0} & 68.6 & 54.3 & \textbf{94.4\%} \\
    \midrule
    \rowcolor{mygray}
    \multicolumn{7}{c}{\textit{Retain 16 Tokens} \ \fg{$(\downarrow 97.2\%)$}} \\
    VisionZip & 47.0 & 1350 & 66.6 & 67.8 & 49.8 & 81.8\% \\
    HoloV     & 48.9 & 1444 & 70.6 & 68.1 & 50.7 & 84.8\% \\
    FlowCut   & 48.2 & 1357 & 71.1 & 67.2 & 47.4 & 82.3\% \\
    PRUNESID  & 50.3 & 1436 & 74.5 & 67.3 & 49.5 & 85.4\% \\
    SCOPE     & 53.4 & 1562 & 76.9 & 68.2 & 52.0 & 89.4\% \\
    \rowcolor{lightgreen!80}
    Ours & \textbf{55.1} & \textbf{1624} & \textbf{81.5} & \textbf{69.3} & \textbf{52.1} & \textbf{92.1\%} \\
    \bottomrule[1.2pt]
  \end{tabular}
  }
\vspace{-10pt}
\end{table}

\begin{table}[t]
  \centering
  \small
  \setlength{\tabcolsep}{5pt}
  \renewcommand{\arraystretch}{1.05}
  \caption{Performance comparison under different visual token settings on LLaVA-NeXT-7B. The vanilla number of visual tokens is 2880, and the last column reports the average accuracy proportion w.r.t. the vanilla.}
  \label{tab:llavanext7b}
  \resizebox{\columnwidth}{!}{
  \begin{tabular}{lcccccc}
    \toprule[1.2pt]
    \textbf{Method} & \textbf{GQA} & \textbf{MME} & \textbf{POPE} & \textbf{SQA} & \textbf{TextVQA} & \textbf{Avg.} \\
    \midrule
    \rowcolor{mygray}
    \multicolumn{7}{c}{\textit{Upper Bound, 2880 Tokens} \ \textbf{(100.0\%)}} \\
    \textcolor{gray}{Vanilla} & \textcolor{gray}{64.2} & \textcolor{gray}{1842} & \textcolor{gray}{86.4} & \textcolor{gray}{70.2} & \textcolor{gray}{61.3} & \textcolor{gray}{100.0\%} \\
    \midrule
    \rowcolor{mygray}
    \multicolumn{7}{c}{\textit{Retain 160 Tokens} \ \fg{$(\downarrow 94.4\%)$}} \\
    VisionZip & 55.5$^{\dagger}$ & 1630$^{\dagger}$ & 84.2\hphantom{$^{\dagger}$} & \textbf{68.3}$^{\dagger}$ & 56.2$^{\dagger}$ & 92.3\% \\
    HoloV     & 57.2\hphantom{$^{\dagger}$} & 1675\hphantom{$^{\dagger}$} & 81.4\hphantom{$^{\dagger}$} & 67.1\hphantom{$^{\dagger}$} & 52.3\hphantom{$^{\dagger}$} & 91.0\% \\
    FlowCut   & 57.6$^{\dagger}$ & 1746$^{\dagger}$ & 79.9$^{\dagger}$ & 66.5\hphantom{$^{\dagger}$} & 57.6$^{\dagger}$ & 93.1\% \\
    PRUNESID  & 58.9$^{\dagger}$ & 1704$^{\dagger}$ & 76.9$^{\dagger}$ & 67.1$^{\dagger}$ & \textbf{59.1}\hphantom{$^{\dagger}$} & 93.1\% \\
    SCOPE     & \textbf{60.0}\hphantom{$^{\dagger}$} & 1700$^{\dagger}$ & 83.5\hphantom{$^{\dagger}$} & 67.4$^{\dagger}$ & 56.8$^{\dagger}$ & 94.2\% \\
    \rowcolor{lightgreen!80}
    Ours & 59.7\hphantom{$^{\dagger}$} & \textbf{1764}\hphantom{$^{\dagger}$} & \textbf{84.6}\hphantom{$^{\dagger}$} & 67.5\hphantom{$^{\dagger}$} & 55.3\hphantom{$^{\dagger}$} & \textbf{94.6\%} \\
    \midrule
    \rowcolor{mygray}
    \multicolumn{7}{c}{\textit{Retain 80 Tokens} \ \fg{$(\downarrow 97.2\%)$}} \\
    VisionZip & 52.3 & 1534 & 75.2 & \textbf{68.3} & 53.4 & 87.2\% \\
    HoloV     & 48.7 & 1432 & 69.1 & 67.7 & 46.9 & 81.3\% \\
    FlowCut   & 54.9 & 1550 & 78.5 & \textbf{68.3} & \textbf{55.2} & 89.6\% \\
    PRUNESID  & 56.3 & 1642 & 78.9 & 66.8 & 49.9 & 88.9\% \\
    SCOPE     & 57.6 & 1636 & 79.1 & 68.2 & 53.0 & 90.7\% \\
    \rowcolor{lightgreen!80}
    Ours & \textbf{58.7} & \textbf{1692} & \textbf{80.8} & 67.1 & 53.3 & \textbf{91.9\%} \\
    \midrule
    \rowcolor{mygray}
    \multicolumn{7}{c}{\textit{Retain 40 Tokens} \ \fg{$(\downarrow 98.6\%)$}} \\
    VisionZip & 44.1 & 1150 & 62.4 & 65.5 & 47.0 & 74.7\% \\
    HoloV     & 44.3 & 1229 & 62.2 & 66.5 & 43.4 & 74.6\% \\
    FlowCut   & 50.7 & 1401 & 72.6 & \textbf{68.7} & \textbf{50.3} & 83.8\% \\
    PRUNESID  & 49.4 & 1261 & 66.1 & 67.5 & 45.6 & 78.5\% \\
    SCOPE     & 54.3 & 1515 & 73.0 & 68.4 & 48.4 & 85.5\% \\
    \rowcolor{lightgreen!80}
    Ours & 56.1 & \textbf{1655} & \textbf{76.5} & 66.7 & 50.0 & \textbf{88.5\%} \\
    \midrule
    \rowcolor{mygray}
    \multicolumn{7}{c}{\textit{Retain 20 Tokens} \ \fg{$(\downarrow 99.3\%)$}} \\
    VisionZip & 42.3 & 1134 & 65.2 & 67.3 & 43.4 & 73.9\% \\
    HoloV     & 41.2 & 1051 & 56.2 & 64.5 & 41.1 & 69.0\% \\
    FlowCut   & 40.6 & 1012 & 53.6 & 65.2 & 41.0 & 68.0\% \\
    PRUNESID  & 44.8 & 1002 & 54.3 & 67.0 & 39.0 & 69.2\% \\
    SCOPE     & 47.8 & 1298 & 60.4 & 67.3 & 42.4 & 76.0\% \\
    \rowcolor{lightgreen!80}
    Ours & \textbf{51.9} & \textbf{1379} & \textbf{68.6} & \textbf{68.0} & \textbf{43.9} & \textbf{80.7\%} \\
    \bottomrule[1.2pt]
  \end{tabular}
  }
\vspace{-10pt}
\end{table}

\paragraph{Results on LLaVA-NeXT.}

We further evaluate our method on LLaVA-NeXT-7B, which contains up to 2880 visual tokens, posing a higher risk of feature distribution inconsistency under ultra-low token pruning.

As shown in Table~\ref{tab:llavanext7b}, our method also achieves strong performance on LLaVA-NeXT-7B, which contains substantially more visual tokens. Under moderate pruning with 160 retained tokens, different methods achieve relatively similar performance, while our method still slightly outperforms SCOPE in average accuracy, demonstrating that our method remains effective even when the original model contains substantially more visual tokens.

The advantage of our method becomes increasingly pronounced as the token budget decreases. At 80 and 40 retained tokens, our method surpasses SCOPE by 1.2\% and 3.0\% in average accuracy, respectively. Under the most aggressive setting with only 20 retained tokens, our method still preserves 80.7\% of the upper-bound performance and exceeds all baselines by at least 4.7\%, while competing methods are at or below 76.0\%.

The gains are particularly evident on GQA, MME, and POPE. Under the 20 token setting, our method outperforms SCOPE by 4.1\% on GQA and 8.2\% on POPE. 
These results show that our method maintains stronger performance under extreme compression.

\paragraph{Results on Video-LLaVA.}

We further evaluate our method on Video-LLaVA under aggressive visual token pruning. Because video understanding requires capturing both spatial and temporal dependencies, it is more sensitive to token reduction.

As shown in Table~\ref{tab:videollava7b}, our method achieves the best average performance under both token budgets, retaining 79.52\% and 69.89\% of the vanilla accuracy with only 128 and 64 visual tokens, respectively. In particular, our method shows clear advantages on TGIF, reaching 33.60\% at 128 tokens and 30.57\% at 64 tokens, outperforming SCOPE by 7.85\% and 10.10\%, respectively. It also consistently surpasses FlowCut across all datasets and token budgets. This suggests that even under aggressive pruning, our method preserves most of the original performance. This demonstrates the effectiveness of our framework on video language tasks. These findings indicate that substantial redundancy exists in video inputs, suggesting that aggressive token pruning can significantly improve the efficiency of video LLMs without causing severe performance degradation.

\begin{table}[t]
  \centering
  \small
  \setlength{\tabcolsep}{8.5pt}
  \renewcommand{\arraystretch}{1.2}
  \caption{Performance comparison on Video-LLaVA under different visual token settings. The vanilla number of visual tokens is 2048, and the last column reports the average accuracy proportion w.r.t. the vanilla.}
  \vspace{-5pt}
  \label{tab:videollava7b}
  \resizebox{\columnwidth}{!}{
  \begin{tabular}{ccccc}
    \toprule[1.2pt]
    \textbf{Method} & \textbf{TGIF} & \textbf{MSVD} & \textbf{MSRVTT} & \textbf{Avg.} \\
    \midrule
    \rowcolor{mygray}
    \multicolumn{5}{c}{\textit{Upper Bound, 2048 Tokens} \ \textbf{(100.0\%)}} \\
    \textcolor{gray}{Vanilla} & \textcolor{gray}{45.70} & \textcolor{gray}{68.09} & \textcolor{gray}{54.49} & \textcolor{gray}{100.0\%} \\
    \midrule

    \rowcolor{mygray}
    \multicolumn{5}{c}{\textit{Retain 128 Tokens} \ \fg{$(\downarrow 93.8\%)$}} \\
    FlowCut & 29.39 & 46.16 & 37.45 & 66.95\% \\
    SCOPE   & 25.75 & \textbf{60.06} & 44.81 & 75.59\% \\
    \rowcolor{lightgreen!80}
    Ours    & \textbf{33.60} & 52.76 & \textbf{47.70} & \textbf{79.52\%} \\
    \midrule

    \rowcolor{mygray}
    \multicolumn{5}{c}{\textit{Retain 64 Tokens} \ \fg{$(\downarrow 96.9\%)$}} \\
    FlowCut & 25.80 & 38.38 & 31.39 & 56.82\% \\
    SCOPE   & 20.47 & \textbf{55.55} & \textbf{40.81} & 67.09\% \\
    \rowcolor{lightgreen!80}
    Ours    & \textbf{30.57} & 50.24 & 37.59 & \textbf{69.89\%} \\
    \bottomrule[1.2pt]
  \end{tabular}
  }
\end{table}

\begin{table}[t]
  \centering
  \small
  \setlength{\tabcolsep}{3.5pt}
  \renewcommand{\arraystretch}{1.2}
  \caption{Performance comparison on Qwen2-VL-7B under 95\% token reduction ratio. The last column reports the average accuracy proportion w.r.t. the vanilla.}
  \vspace{-5pt}
  \label{tab:qwen2vl7b}
  \resizebox{\columnwidth}{!}{
  \begin{tabular}{lcccccc}
    \toprule[1.2pt]
    \textbf{Method} & \textbf{GQA} & \textbf{MME} & \textbf{POPE} & \textbf{SQA} & \textbf{TextVQA} & \textbf{Avg.} \\
    \midrule
    \rowcolor{mygray}
    \multicolumn{7}{c}{\textit{Upper Bound} \ \textbf{(100.0\%)}} \\
    \textcolor{gray}{Vanilla} & \textcolor{gray}{61.9} & \textcolor{gray}{2302} & \textcolor{gray}{88.8} & \textcolor{gray}{84.6} & \textcolor{gray}{81.3} & \textcolor{gray}{100.0\%} \\
    \midrule
    \rowcolor{mygray}
    \multicolumn{7}{c}{\textit{Retain 5\% Tokens} \ \fg{$(\downarrow 95.0\%)$}} \\
    FlowCut & 50.7 & 2036 & 79.3 & 78.2 & 62.4 & 85.8\% \\
    SCOPE & 52.1 & 2085 & 81.2 & 78.3 & \textbf{65.2} & 87.8\% \\
    \rowcolor{lightgreen!80}
    Ours & \textbf{54.3} & \textbf{2134} & \textbf{82.7} & \textbf{79.1} & 64.6 & \textbf{89.3\%} \\
    \bottomrule[1.2pt]
  \end{tabular}
  }
\vspace{-10pt}
\end{table}

\paragraph{Results on Qwen2-VL.} 
Table~\ref{tab:qwen2vl7b} presents the performance comparison on Qwen2-VL-7B under a 95\% token reduction ratio. Our method achieves the best average performance of 89.3\%, outperforming SCOPE and FlowCut by 1.5\% and 3.5\%, respectively, further demonstrating its effectiveness under ultra-low token budget. 
Moreover, our method achieves the best performance on GQA, MME, POPE, and SQA under ultra-low token budget, with particularly significant gains on GQA and MME, where it surpasses SCOPE by 2.2\% and 49 points, respectively. These results demonstrate the consistent advantages of our approach across diverse multimodal understanding benchmarks.

\begin{table}[t]
  \centering
  \small
  \setlength{\tabcolsep}{13pt}
  \renewcommand{\arraystretch}{1.08}
  \caption{\textit{Absolute} Pearson correlations between $\kappa$ and accuracy across token budgets from 10 to 100.}
  \vspace{-5pt}
  \label{tab:kl_acc_corr}
  \resizebox{\columnwidth}{!}{
  \begin{tabular}{lccc}
    \toprule
    \textbf{Method} & \textbf{GQA} & \textbf{POPE} & \textbf{TextVQA} \\
    \midrule
    VisionZip & 0.95 & 0.90 & 0.94 \\
    HoloV     & 0.86 & 0.95 & 0.82 \\
    FlowCut   & 0.96 & 0.99 & 0.96 \\
    PRUNESID  & 0.98 & 0.98 & 0.83 \\
    SCOPE     & 0.98 & 0.93 & 0.82 \\
    Ours      & 0.97 & 0.94 & 0.87 \\
    \bottomrule
  \end{tabular}
  }
\end{table}

\begin{table}[t]
  \centering
  \setlength{\tabcolsep}{3.5pt}
  \renewcommand{\arraystretch}{1.12}
  \caption{Effect of ACGR and TATCS on performance and feature distribution consistency under ultra-low token pruning on LLaVA-1.5-7B. Each entry reports the \textit{accuracy / $\kappa$}, where smaller $\kappa$ (Eq.~\ref{eq:tatcs_instability}) indicates better feature distribution consistency.}
  \vspace{-5pt}
  \label{tab:ablation_llava1.5_7b}
  \resizebox{\columnwidth}{!}{
  \begin{tabular}{ccccc}
  
    \toprule[1.2pt]
    {\textbf{Dataset}} &
    {\textbf{Tokens}} &
    {\textbf{w/o TATCS}} &
    {\textbf{w/o ACGR}} &
    {\textbf{ACGR+TATCS}} \\
    \midrule

    \multirow{3}{*}{GQA}
    & 16 & 37.8 / 1.06 & 50.5 / 0.78 & \textbf{55.1 / 0.66} \\
    & 32 & 37.5 / 0.42 & 54.1 / 0.34 & \textbf{57.0 / 0.21} \\
    & 64 & 37.7 / 0.43 & 56.7 / 0.15 & \textbf{58.5 / 0.07} \\
    \midrule

    \multirow{3}{*}{POPE}
    & 16 & 50.2 / 1.07 & 72.9 / 0.85 & \textbf{81.5 / 0.71} \\
    & 32 & 49.5 / 0.56 & 79.1 / 0.49 & \textbf{85.0 / 0.23} \\
    & 64 & 50.2 / 0.46 & 82.6 / 0.10 & \textbf{86.5 / 0.07} \\

    \bottomrule[1.2pt]
  \end{tabular}
  }
\vspace{-10pt}
\end{table}

\subsection{Distribution Consistency and Ablation Analysis}

\paragraph{Distribution Consistency and Performance Correlation.}

{
As illustrated in Fig.~\ref{fig:ackl}(a), reducing the token budget progressively increases the distribution deviation $\kappa$, while the downstream accuracy consistently decreases. Under moderate token budgets, both $\kappa$ and accuracy change relatively smoothly. However, under ultra-low token budget, $\kappa$ increases more rapidly and is accompanied by substantially larger performance degradation as the pruning budget is further reduced. This consistent trend indicates that larger distribution deviation is generally associated with lower accuracy under ultra-low token budget.
}

To further examine this relationship, Table~\ref{tab:kl_acc_corr} reports the absolute Pearson correlation coefficient between $\kappa$ and downstream accuracy across token budgets ranging from 10 to 100. Strong correlations are consistently observed across different benchmarks and pruning methods. In particular, the correlations on GQA remain above 0.86 for all methods, while POPE consistently achieves values at or above 0.90. TextVQA also exhibits high correlations ranging from 0.82 to 0.96. These results show that, under ultra-low token budget, the strong correlation between distribution deviation and accuracy exists across different pruning strategies and benchmarks. Therefore, $\kappa$ can serve as an effective lightweight indicator for analyzing the behavior of aggressive compression.

\paragraph{Ablation Analysis on ACGR and TATCS.}
{
Removing ACGR consistently degrades performance, with the degradation becoming more severe as the token budget decreases, demonstrating the importance of ACGR during progressive pruning. As shown in Table~\ref{tab:ablation_llava1.5_7b}, removing ACGR on POPE leads to accuracy drops of 3.9\%, 5.9\%, and 8.6\% at token budgets of 64, 32, and 16, respectively. The progressively larger performance gap suggests that ACGR provides greater benefits under more aggressive pruning settings. These results indicate that enriching retained tokens with contextual information before token removal helps alleviate performance degradation under ultra-low token budget.
}

{
Removing TATCS causes substantial performance degradation across both datasets, highlighting its importance under ultra-low token pruning. As shown in Table~\ref{tab:ablation_llava1.5_7b}, without TATCS, the accuracy on GQA remains around 37.5\%-37.8\% even as the token budget increases from 16 to 64, while a similar trend is observed on POPE, where the accuracy stays around 50\% across different token budgets. Meanwhile, $\kappa$ remains consistently high without TATCS. These results suggest that simply increasing the number of retained tokens does not necessarily improve performance under ultra-low token pruning. In contrast, TATCS substantially reduces $\kappa$ and consistently improves accuracy across different token budgets, indicating its effectiveness under ultra-low token budget.
}

\begin{table}[t]
  \centering
  \small
  \setlength{\tabcolsep}{5pt}
  \renewcommand{\arraystretch}{1.08}
  \caption{Efficiency comparison on POPE under the 32 tokens setting. \(\Delta\) is the speed-up computed w.r.t. the vanilla. Our method clearly outperforms existing methods while maintaining comparable efficiency.}
  \vspace{-5pt}
  \label{tab:efficiency_analysis}
  \resizebox{\columnwidth}{!}{
  \begin{tabular}{lcccc}
    \toprule[1.2pt]
    \textbf{Method} & \textbf{Tokens} & \textbf{Accuracy} & \textbf{Time (s)} & \textbf{\(\Delta\)} \\
    \midrule
    \textcolor{gray}{Vanilla} & \textcolor{gray}{576} & \textcolor{gray}{85.9} & \textcolor{gray}{2147} & \textcolor{gray}{\(1.0\times\)} \\  \midrule
    VisionZip & 32 & 74.7 & 901 & \(2.4\times\) \\
    HoloV & 32 & 76.1 & 876 & \(\textbf{2.5}\times\) \\
    FlowCut & 32 & 75.5 & \textbf{856} & \(\textbf{2.5}\times\) \\
    PRUNESID & 32 & 82.1 & 930 & \(2.3\times\) \\
    SCOPE & 32 & 82.4 & 884 & \(2.4\times\) \\
    Ours & 32 & \textbf{85.0} & 895 & \(2.4\times\) \\
    \bottomrule[1.2pt]
  \end{tabular}
  }
\end{table}

\begin{table}[t]
\centering
\small
\setlength{\tabcolsep}{5pt}
\renewcommand{\arraystretch}{1.08}
\caption{
Performance drop after removing each component under ultra-low token pruning on LLaVA-1.5-7B.
Rows "w/o $s_i^{\mathrm{cos}}$", "w/o $s_i^{\mathrm{var}}$", and "w/o $s_i^{\mathrm{abl}}$" remove the corresponding ACGR score terms in Eq.~\ref{acgr:score}; "w/o Text" removes the text-aware term of TATCS in Eq.~\ref{tatcs:text}.
}
\vspace{-5pt}
\label{tab:component_ablation}
\resizebox{\columnwidth}{!}{
\begin{tabular}{lcccccc}
\toprule[1.2pt]
& \multicolumn{3}{c}{GQA} 
& \multicolumn{3}{c}{POPE} \\
\cmidrule(lr){2-4}
\cmidrule(lr){5-7}
\multirow{2}{*}[3.4ex]{Component}
& 64 & 32 & 16
& 64 & 32 & 16 \\
\midrule

w/o $s_i^{\mathrm{cos}}$
& -0.2 & -0.6 & -0.3
& -0.7 & -1.1 & -0.4 \\

w/o $s_i^{\mathrm{var}}$
& -0.4 & -0.3 & -0.5
& -0.4 & -0.2 & -0.1 \\

w/o $s_i^{\mathrm{abl}}$
& -0.3 & -0.1 & -0.8
& -0.7 & -0.9 & -0.5 \\

w/o Text
& -0.2 & -0.2 & -0.3
& -0.5 & -0.7 & -0.5 \\

\bottomrule[1.2pt]
\end{tabular}
}
\vspace{-10pt}
\end{table}

\subsection{Component and Efficiency Analysis}

Table~\ref{tab:efficiency_analysis} compares our method with existing pruning approaches under the same 32 token budget. All pruning methods achieve about 2.3$\times$ to 2.5$\times$ acceleration over the Vanilla model, demonstrating the efficiency benefits of aggressive token reduction. Among them, our method achieves the highest accuracy of 85.0\%, outperforming the strongest baseline SCOPE by 2.6\% while maintaining a comparable inference time of 895s versus 884s. Although FlowCut is slightly faster, it suffers from noticeable accuracy degradation under the same ultra-low token budget. These results show that our method achieves a better balance between accuracy and efficiency while maintaining comparable acceleration.

\paragraph{Effect of Individual Components.}

{
Table~\ref{tab:component_ablation} shows that different ACGR score terms play complementary roles under ultra-low token pruning. Specifically, $s_i^{\mathrm{cos}}$ and $s_i^{\mathrm{var}}$ complement the $s_i^{\mathrm{cls}}$ score by modeling feature redundancy and transformation dynamics, respectively. In addition, $s_i^{\mathrm{abl}}$ further incorporates token influence through attention-ablation estimation, providing an additional importance signal beyond individual token saliency. This becomes increasingly beneficial under more aggressive pruning, where tokens with larger influence on other tokens may not always be assigned high importance by standard saliency-based estimation.
}

{
Removing the text-aware term in TATCS consistently degrades performance across various token budgets and benchmarks. As shown in Table~\ref{tab:component_ablation}, its removal causes performance drops on both GQA and POPE, with a more noticeable 0.7\% decline on POPE at 32 tokens. This suggests that text-aware guidance becomes particularly important for hallucination-sensitive benchmarks such as POPE, where preserving text-relevant visual evidence is more critical for accurate prediction. Without text-aware guidance, visually salient but text-irrelevant tokens may dominate the clustering process, losing important visual information and thereby degrading accuracy under ultra-low token budget.
}

\section{Conclusion}
In this paper, we investigate ultra-low visual token pruning from the perspective of feature distribution consistency and show that aggressively reducing the token budget induces significant distribution shift in kept token representations, accompanied by performance degradation. 
To address this issue, we propose ACGR to transfer complementary contextual information from pruned tokens to retained anchor tokens before removal, and TATCS to mitigate distribution shift through text-aware representative token reselection. 
Extensive experiments on multiple vision-language models demonstrate that our method consistently achieves superior performance under ultra-low token budget, retaining 92.1\% of the upper-bound performance on LLaVA-1.5-7B using only 16 visual tokens. 
Overall, our findings highlight the importance of preserving feature distribution consistency for ultra-low visual token pruning.

\section*{Limitations}
Although the proposed framework achieves strong performance under ultra-low token pruning, a noticeable gap still exists compared with using full visual tokens. Similar to existing pruning strategies, aggressive token compression inevitably causes partial information loss, making it difficult to fully preserve visual semantics under ultra-low token budget. We will explore more effective semantic preservation and adaptive token allocation strategies to further improve the balance between efficiency and performance.

\bibliography{custom}

\newpage

\appendix

\section{Detailed Definitions of Token Importance Scores}
\label{sec:score}
To better assess token importance under ultra-low token budget, we combine multiple complementary criteria from different perspectives, including semantic relevance, redundancy suppression, feature transformation dynamics, and attention interaction effects. 
Given the patch tokens in the current layer \(X=\{x_i\}_{i=1}^{M}\), the overall importance score of token \(x_i\) is defined as
\begin{equation}
s_i = s_i^{\mathrm{cls}} + s_i^{\mathrm{cos}} + s_i^{\mathrm{var}} + s_i^{\mathrm{abl}} .
\end{equation}

\paragraph{CLS relevance score.}
To estimate the semantic importance of visual tokens, we introduce a CLS-guided relevance score based on the attention from the CLS token to each image patch. Tokens receiving higher CLS attention are regarded as more semantically informative and are thus more likely to be preserved during pruning \citep{zhang2025beyond, liang2022not, long2023beyond}. 
Let \(A_{0,i}^{(h)}\) denote the attention weight from the CLS token to patch \(i\) at head \(h\), where \(H\) is the number of attention heads. We first average the attention weights across heads and then normalize them across all tokens:
\begin{equation}
\alpha_i^{\mathrm{cls}}=\frac{1}{H}\sum_{h=1}^{H}A_{0,i}^{(h)}.
\end{equation}
\begin{equation}
s_i^{\mathrm{cls}}=\frac{\alpha_i^{\mathrm{cls}}}{\sum_{j=1}^{M}\alpha_j^{\mathrm{cls}}+\varepsilon}.
\end{equation}

\paragraph{Cosine non-redundancy score.}
To suppress redundancy among visual tokens, we measure the average cosine similarity between each token and all remaining tokens. Tokens that are less similar to others are considered to contain more unique information and are therefore assigned higher non-redundancy scores \citep{bolya2022token, jeddi2025similarity, wang2024zero}. 
Specifically, we compute the average cosine similarity of each token to all other tokens and transform it into a non-redundancy score:
\begin{equation}
c_i=1-\frac{1}{M-1}\sum_{\substack{j=1\\j\neq i}}^{M}\cos(x_i,x_j).
\end{equation}
\begin{equation}
s_i^{\mathrm{cos}}=\frac{c_i}{\sum_{j=1}^{M}c_j+\varepsilon}.
\end{equation}

\paragraph{Transformation variation score.}
Tokens exhibiting larger feature updates through the attention and FFN branches are considered to encode richer semantic information and are therefore assigned higher importance scores \citep{ishibashi2025automatic, wu2024token}.
This component follows the transition-based perspective in TransPrune \citep{li2025transprune}.
Let \(x_i^{\mathrm{in}}\), \(x_i^{\mathrm{att}}\), and \(x_i^{\mathrm{ffn}}\) denote the patch feature at the layer input, after the attention branch, and after the FFN branch in the same layer, respectively.
We first measure the transformation magnitude of token \(i\) in the attention and FFN branches by cosine distance:
\begin{equation}
d_i^{\mathrm{att}}=1-\cos(x_i^{\mathrm{in}},x_i^{\mathrm{att}}),
\quad
d_i^{\mathrm{ffn}}=1-\cos(x_i^{\mathrm{in}},x_i^{\mathrm{ffn}}).
\end{equation}
The distances are then normalized across all patch tokens with a softmax operation:
\begin{equation}
p_i^{\mathrm{att}}=\frac{\exp(d_i^{\mathrm{att}})}{\sum_{j=1}^{M}\exp(d_j^{\mathrm{att}})},
\quad
p_i^{\mathrm{ffn}}=\frac{\exp(d_i^{\mathrm{ffn}})}{\sum_{j=1}^{M}\exp(d_j^{\mathrm{ffn}})}.
\end{equation}
To further account for the change in feature magnitude, we compute the norm ratios:
\begin{equation}
r_i^{\mathrm{att}}=\frac{\lVert x_i^{\mathrm{att}}\rVert_2}{\lVert x_i^{\mathrm{in}}\rVert_2+\varepsilon},
\quad
r_i^{\mathrm{ffn}}=\frac{\lVert x_i^{\mathrm{ffn}}\rVert_2}{\lVert x_i^{\mathrm{in}}\rVert_2+\varepsilon}.
\end{equation}
The final transformation variation score is computed by combining the normalized transformation weights and magnitude ratios:
\begin{equation}
s_i^{\mathrm{var}}=p_i^{\mathrm{att}}r_i^{\mathrm{att}}+p_i^{\mathrm{ffn}}r_i^{\mathrm{ffn}}.
\end{equation}

\paragraph{Attention-ablation impact score.}
Some tokens may appear locally unimportant while still exerting strong influence on other token representations through Transformer attention interactions \citep{chefer2021transformer, sundararajan2017axiomatic}. To capture this effect, we estimate token importance by measuring how much the representations of other tokens change after removing a target token from the attention process. Tokens inducing larger representation variations are considered more important for contextual information propagation.

For a fixed attention head, the attention output of token \(i\) is
\begin{equation}
a_i=\sum_{j=1}^{M}\alpha_{ij}v_j,
\end{equation}
where \(M\) is the number of patch tokens, \(\alpha_{ij}\) is the attention weight from token \(i\) to token \(j\), and \(v_j\) is the value vector of token \(j\).

If token \(k\) is removed, the remaining attention weights are renormalized as
\begin{equation}
\tilde{a}_i^{(-k)}
=
\sum_{j\neq k}\frac{\alpha_{ij}}{1-\alpha_{ik}+\varepsilon}v_j
=
\frac{a_i-\alpha_{ik}v_k}{1-\alpha_{ik}+\varepsilon},
\end{equation}
where \(\varepsilon\) is a small constant for numerical stability. The resulting output change is approximated by
\begin{equation}
\Delta a_i^{(k)}
=
\tilde{a}_i^{(-k)}-a_i
\approx
\frac{\alpha_{ik}}{1-\alpha_{ik}+\varepsilon}(a_i-v_k).
\end{equation}
We estimate the influence of token \(k\) on token \(i\) as
\begin{equation}
\hat{I}_{k\rightarrow i}
=
\frac{\alpha_{ik}}{1-\alpha_{ik}+\varepsilon}
\lVert a_i-v_k\rVert_2.
\end{equation}

To avoid re-running the Transformer for each removed token, we further use a one-pass proxy:
\begin{equation}
I_{k\rightarrow i}
=
\frac{\bar{\alpha}_{ik}}{1-\bar{\alpha}_{ik}+\varepsilon}
\cdot
\lVert y_i\rVert_2
\cdot
\lVert \bar{v}_k\rVert_2,
\end{equation}
where \(H\) is the number of attention heads, \(\bar{\alpha}_{ik}\) is the head-averaged attention weight,
$\bar{v}_k=\frac{1}{H}\sum_{h=1}^{H}v_{h,k}$,
and \(y_i\) is the current-layer hidden feature of token \(i\) obtained from the same forward pass.

Finally, the attention-ablation importance score is computed as
\begin{equation}
s_k^{\mathrm{abl}}
=
\frac{1}{M-1}
\sum_{\substack{i=1 \\ i\neq k}}^{M}
I_{k\rightarrow i}.
\end{equation}

\section{Distribution Consistency Metric}
\label{app:gaussian_kappa}

Existing pruning strategies are generally importance-oriented or coverage-oriented, yet both tend to become unstable under ultra-low token budget \citep{fayyaz2022adaptive}. As the token budget decreases, retained tokens progressively deviate from the original feature distribution \citep{rao2021dynamicvit}, leading to degraded multimodal reasoning \citep{ioffe2015batch}. Motivated by this observation, we revisit ultra-low token pruning from the perspective of distribution consistency between retained and full token representations.

A common way to measure distribution discrepancy is through divergence metrics such as KL divergence or Wasserstein distance \citep{dowson1982frechet,sun2016deep}. 
However, these methods typically require covariance inversion, eigendecomposition, or determinant computation, leading to considerable computational overhead and numerical instability for runtime token pruning \citep{arsigny2006log}. 
Instead of estimating exact probability divergence, we adopt a lightweight distribution consistency approximation based on first- and second-order feature statistics:
\begin{equation}
\begin{aligned}
\mathcal{D}(X_{k},X_{f})&=
\frac{
\mathrm{Tr}(\Sigma_{k})+d\,\mathrm{Var}(\mu_{k})
}{
\mathrm{Tr}(\Sigma_{f})+d\,\mathrm{Var}(\mu_{f})
}, \\
\kappa&=|\mathcal{D}(X_{k},X_{f})-1|,
\end{aligned}
\end{equation}
where $X_f$ and $X_k$ denote the full token and kept token representations, respectively.

Here, $\mathrm{Tr}(\Sigma)$ denotes the trace of the covariance matrix and serves as a lightweight approximation of second-order feature statistics, reflecting the overall feature dispersion \citep{sun2017correlation,li2018towards}, while $\mathrm{Var}(\mu)$ measures the variance of channel-wise mean activations and reflects shifts in channel-wise feature responses \citep{huang2017arbitrary,ulyanov2016instance,adachi2023covariance}. Since $\mathrm{Tr}(\Sigma)$ accumulates variance across feature dimensions, multiplying $\mathrm{Var}(\mu)$ by $d$ balances the scales of first- and second-order statistics.

The ratio formulation provides a scale-normalized comparison between kept and full token representations. When the kept tokens preserve distribution consistency with the original tokens representation, $\mathcal{D}(X_k,X_f)$ approaches $1$, yielding a smaller $\kappa$.
Therefore, $\kappa$ is not intended as a strict probability divergence metric, but rather as a lightweight proxy for distribution deviation under ultra-low token pruning.

Although $\kappa$ does not explicitly characterize representation geometry or fine-grained semantic structure, it provides an efficient empirical indicator of whether aggressive pruning introduces substantial statistical distribution deviation during inference. 
As shown in Table~\ref{tab:kl_acc_corr}, larger $\kappa$ consistently correlates with more severe downstream accuracy degradation under ultra-low token budget. 
Motivated by this observation, we use $\kappa$ as an adaptive trigger signal for TATCS, which is activated only when the estimated distribution deviation exceeds a predefined threshold, thereby dynamically mitigating distribution shift in the retained token features.

\section{Visualization of Kept Tokens}
\begin{figure*}[t]
  \centering
  \includegraphics[width=0.75\textwidth]{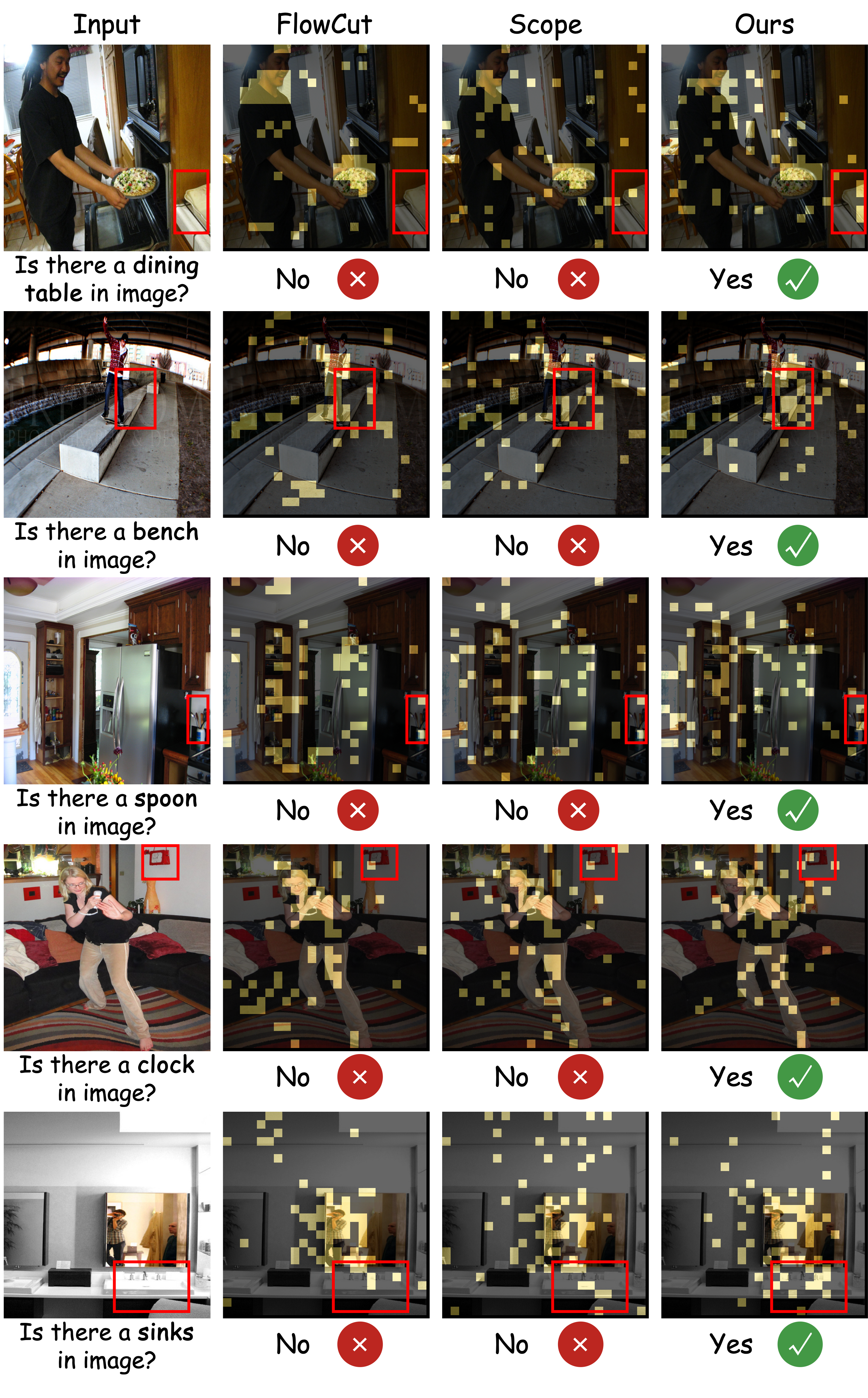}
  \caption{Qualitative visualization of selected visual tokens under a token budget of 64 on LLaVA-1.5-7B.}
  \label{fig:vis1}
\end{figure*}

\begin{figure*}[t]
  \centering
  \includegraphics[width=0.75\textwidth]{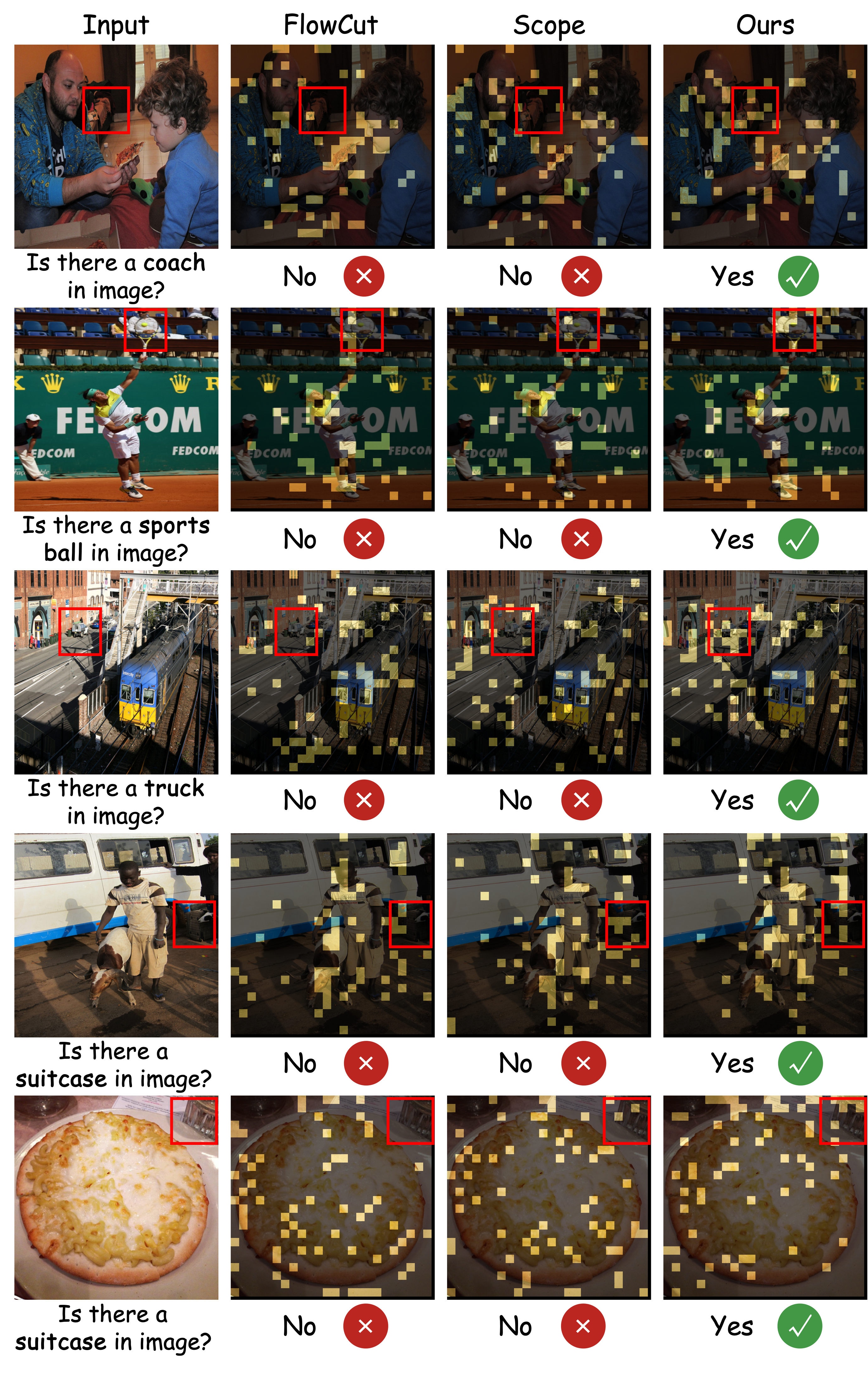}
  \caption{Qualitative visualization of selected visual tokens under a token budget of 64 on LLaVA-1.5-7B.}
  \label{fig:vis2}
\end{figure*}

Fig.~\ref{fig:vis1} and Fig.~\ref{fig:vis2} compare token selection under 64 token budget on LLaVA-1.5-7B. FlowCut focuses on salient local regions, while SCOPE favors spatial coverage. By maintaining feature distribution consistency, our method preserves both semantic and contextual information, resulting in a more representative token subset.

\end{document}